\documentclass[11pt]{article}

\usepackage[T1]{fontenc}
\usepackage{lmodern}
\usepackage[letterpaper,margin=1in]{geometry}
\usepackage{microtype}
\usepackage{amsmath,amssymb}
\usepackage{graphicx}
\usepackage{float}
\usepackage[font=small,labelfont=bf,skip=6pt]{caption}
\usepackage{xcolor}
\usepackage[numbers,sort&compress]{natbib}
\usepackage{fancyhdr}
\usepackage{hyperref}

\hypersetup{
  colorlinks=true,
  linkcolor=blue!45!black,
  citecolor=blue!45!black,
  urlcolor=blue!55!black,
  pdftitle={VideoGen Model 1: Prompt Expansion Versus Formatting in Audiovisual Generation},
  pdfauthor={David Grossman and Anton Koenig}
}

\fancypagestyle{plain}{%
  \fancyhf{}%
  \fancyfoot[C]{\thepage}%
}

\title{VideoGen Model 1:\\Prompt Expansion Versus Formatting\\in Audiovisual Generation}

\author{%
  David Grossman\\
  CTO, \texorpdfstring{\href{https://videogen.io}{VideoGen}}{VideoGen}\\
  \href{mailto:david@videogen.io}{\texttt{david@videogen.io}}
  \and
  Anton Koenig\\
  CEO, \texorpdfstring{\href{https://videogen.io}{VideoGen}}{VideoGen}\\
  \href{mailto:anton@videogen.io}{\texttt{anton@videogen.io}}%
}
\date{}

\begin{document}
\maketitle
\vspace{-2.5em}

\begin{abstract}
Creative prompt expansion improved overall audiovisual quality scores over raw prompts in a controlled pilot of \href{https://videogen.io/model-1}{VideoGen Model 1}, our post-trained version of MiniMax H3. We compared the raw, reformatted, and expanded versions of 30 prompts, holding the generator, seed, and sampling settings fixed. Each clip was rated separately on five dimensions. Their mean formed an overall score on a 1--10 scale. Compared with formatting alone, overall scores increased by 0.63 points (95\% interval: $[+0.11, +1.15]$) under VA-Judger and 0.15 points (95\% interval: $[-0.03, +0.35]$) under Gemini 3.8 Flash, whose overall change remained uncertain. Prompt-alignment scores increased by 0.97 points under VA-Judger and 0.50 under Gemini. VA-Judger also showed higher audio quality; visual-quality ratings changed little under either judge.
\end{abstract}

\section{Introduction}

Formatted prompts organize the original prompt into H3's supported visual and audio fields (e.g., \texttt{integrated\_\allowbreak{}multimodal\_\allowbreak{}description}, \texttt{overall\_\allowbreak{}soundscape}, and \texttt{non\_\allowbreak{}diegetic\_\allowbreak{}music}). Expanded prompts use the same structure and add creative details about the setting, action, camera movement, performance, and sound.

MiniMaxAI's H3 documentation recommends refining inputs before generation~\cite{minimaxh3}. We tested the added value of creative expansion over a formatting control, using the same VideoGen Model 1 checkpoint in all three conditions.

Prompt optimization for video generation is an established research direction. Cheng et al.~\cite{cheng2025vpo} introduced VPO, which learns to rewrite video prompts through supervised training and preference learning from text- and video-level feedback. Our study examines an instruction-based rewriter at inference time, separating creative expansion from field formatting and evaluating the resulting clips with audio.

\section{Writing architecture}

Video prompts can leave important creative choices unspecified: what happens, how a scene progresses, and what characters say. Our ``writers' room'' is a group of language models assigned to writing, criticism, and reader roles. It develops and revises scripts before a separate clip rewriter translates the material into H3's audiovisual prompt format. This division lets the writers' room focus on story and dialogue while the rewriter handles model-specific generation instructions. Wu et al.~\cite{wu2025movieagent} used hierarchical scene and shot planning in MovieAgent. We evaluate the clip rewriter in Section~\ref{sec:methods}.

\subsection{Writers' room}

\paragraph{Intake and fidelity.}
A structured intake identifies the category (film, marketing, social, explainer, music video, or microdrama), visual medium, aspect ratio, and whether the prompt specifies what should be seen and heard. A detailed text-only prompt that fits one clip goes directly to the rewriter. Longer requests and those with reference images enter the writers' room. Detailed prompts skip concept and story-arc development. The writers' room preserves their requested shots, actions, lines, sounds, style, and order. Open-ended prompts permit creative development within the requested duration and format.

\paragraph{Concepts and scripts.}
For an open-ended prompt, the writers' room proposes six concepts, then five more per subsequent round, over two or three rounds. The writer can consult a reference library, randomly selected creative constraints, and web research. A separate critic applies a category-specific rubric: product relevance and a clear payoff for marketing, or causal progression and an earned ending for film. Rubrics penalize stock story or advertising clich\'es. Marketing and explainer checks also reject invented claims. The two highest-scoring concepts advance. For videos longer than 15 seconds, an arc stage specifies character goals, obstacles, stakes, and timed beats. An arc critic can request one revision. Each finalist becomes a timed screenplay with shared descriptions of character appearances, voices, locations, props, medium, and style.

The primary writer is Claude Opus 5.5, with GPT-6.1 as the critic. Claude Sonnet 5.5 handles intake and reader checks. Later script rounds compare a revision, a bolder alternative, and the previous best. Candidate labels are randomized before criticism. The duration sets a limit of two to five rounds, with early stopping when the critic is satisfied and scores the selected draft at least 8.5/10. These are internal selection scores, not measurements of rendered-video quality. Madaan et al.~\cite{madaan2023selfrefine} proposed Self-Refine, in which a single model generates feedback and revises its own output; our writers' room assigns writing and criticism to separate models.

\paragraph{Clarity and structural checks.}
Three separate model reader calls inspect spoken lines and on-screen text without the explanatory action prose. They retell what they understood and flag confusing or unnatural lines by stable identifiers. Agreement from at least two readers enters the feedback. This targets dialogue that depends on backstory the viewer never hears. Up to two final line-edit passes are retained only if a fresh read flags fewer lines without increasing the number judged confusing. These readers are automated proxies, not human participants.

Code also validates the screenplay's structure. Scene durations must sum to the requested duration. Cast, location, and prop references must resolve. Speakers must belong to the scene or its voiceover and have voice descriptions. The visual medium must match intake. Dialogue has a duration-based word budget with a silent tail reserved before each render's cut. Music cues are checked against the scene and beat structure.

\subsection{Clip-level expansion}

The text-to-video prompt rewriter uses Claude Sonnet 5.5 in three stages: research planning, optional research, and prompt writing. The planner returns 0--2 queries only when a named real product, brand, place, event, or landmark requires visual facts. Research condenses search results into three to eight facts about appearance, material, scale, setting, movement, or sound. Research instructions exclude prices, promotional claims, and history. Generic subjects proceed directly to writing.

The writer receives the prompt, duration, frame dimensions, research notes when available, and H3's prompt guide. It emits three fields: \texttt{integrated\_multimodal\_description} for ordered shots, action, and dialogue; \texttt{overall\_soundscape} for ambience and action-linked sound; and \texttt{non\_diegetic\_music} for background score. The text-to-video instructions preserve supplied dialogue and on-screen text verbatim, prohibit invented dialogue and unrequested real brands, and retain requested framing and shot order. Performance direction describes face, breath, and vocal delivery outside the dialogue tags. Semantic fidelity is governed by these instructions, and code checks the output's structure.

The rewriter targets fewer than 280 words, with a code-enforced ceiling of 320. Validation checks the opening field and shot marker, field order and separators, prohibited wrappers, sequential shot numbering, a duration-based shot-count bound, increasing cut timestamps, and equal counts of opening and closing dialogue tags. Every shot after the first needs a timestamp, and cuts must occur before $\lfloor T-1\rfloor$ seconds for a clip of duration $T$. A failed draft receives one retry with the detected errors. If it fails again, the system returns the original prompt. When music will be composed separately, it places that prompt in the required fields and marks background music absent. The system records research, validation outcomes, and model charges.

Figure~\ref{fig:prompt-example} shows the original prompt and both complete rewrites for one example.

\begin{figure}[htbp]
\centering
\begingroup
\setlength{\fboxsep}{8pt}
\fcolorbox{black!25}{black!2}{%
\begin{minipage}{\dimexpr\linewidth-2\fboxsep-2\fboxrule\relax}
\small
\raggedright
\textbf{Raw}\par
an ad for my soy candle brand, Ember \& Oak\par
\medskip
\textbf{Format-only}\par
\noindent\texttt{integrated\_multimodal\_description}: [Shot 1] An ad for my soy candle brand, Ember \& Oak.\par
\smallskip
\noindent\texttt{overall\_soundscape}: Natural sound matching the action.\par
\smallskip
\noindent\texttt{non\_diegetic\_music}: N/A\par
\medskip
\textbf{Expanded}\par
\noindent\texttt{integrated\_multimodal\_description}: [Shot 1] Live-action, cinematic, horizontal 16:9, a medium shot frames a hand-poured soy candle in an amber glass jar on a cream linen runner over a wooden table, beside a small rust-colored ceramic dish. A plain cream wrap-around paper label on the jar carries the serif text "Ember \& Oak". Soft window light falls from the left as the lit flame sways gently and throws warm reflections across the glass. The camera pushes in with small amplitude at slow speed. [Shot 2] At 00:02.800, the camera cuts to a close-up of the flame above the opaque, creamy off-white wax, where a glossy melted pool glistens and the amber glass glows behind it. The flame stretches and settles as a thin wisp of smoke curls upward. The camera holds a static shot.\par
\smallskip
\noindent\texttt{overall\_soundscape}: Quiet room tone hums softly under the faint flutter of the flame. A gentle breath of air stirs the flame, and the jar glass ticks faintly as it warms.\par
\smallskip
\noindent\texttt{non\_diegetic\_music}: Soft fingerpicked acoustic guitar at a slow tempo over a warm, low synth pad, rising slightly in volume before a gentle fade at the end.\par
\end{minipage}%
}
\endgroup
\caption{Raw, format-only, and expanded prompts for the Ember \& Oak candle ad. All prompt text is reproduced verbatim.}
\label{fig:prompt-example}
\end{figure}

\subsection{Continuity across renders}

For screenplays developed by the writers' room, a director creates shot plans with explicit opening states, spatial blocking, visible characters, dialogue, and timing. A writer uses that plan to compile each clip into the same three-field format. Additional text checks require the planned shot count, opening style sentence, character appearance descriptions, exact dialogue, and exact on-screen text. After two failed writing attempts, code assembles the prompt directly from the plan's fields. Instructions also carry stable speaker identities, off-screen narration, performance, and action-linked sound into each prompt. In this path, background music is composed after the clips are cut together. Clip prompts request scene sound without a background score.

\section{Methods}
\label{sec:methods}

\paragraph{Prompts and generation.}
We authored 30 prompts, three each for commercial content, people and performance, physical action, narrative, nature, animation, fantasy, on-screen text, music, and everyday scenes. Each prompt had one fixed output size: 27 used landscape ($832\times480$), and three used portrait ($480\times832$). We rendered the raw, format-only, and expanded versions of each prompt once at its assigned size, producing $30\times3=90$ clips. All clips used VideoGen Model 1's Diffusers rendering backend, seed 1001, 28 sampling steps, and a B200 GPU. Each clip lasted approximately 5.2 seconds (124 frames at 24~fps).

\paragraph{Prompt conditions.}
\textbf{Raw} used the original prompt verbatim. \textbf{Format-only} used Claude Sonnet 5.5 to organize the prompt into H3's visual/dialogue, soundscape, and background-music fields without creative visual expansion. When sound was unspecified, it inserted ``Natural sound matching the action''. Unspecified music was marked absent. The control therefore included audio defaults as well as formatting. \textbf{Expanded} used the rewriter described above, with research requested for four prompts. The average prompt length was 15 words for raw prompts, 27 for format-only prompts, and 237 for expanded prompts. All rewrites passed structural validation on their first attempt and were used as generated. Background-music instructions were present in 22 expanded prompts and one format-only prompt. These instructions remained in the generation inputs. Expansion cost an average of \$0.02 per prompt in model charges, excluding rendering and evaluation.

\paragraph{Individual-clip evaluation.}
VA-Judger and Gemini 3.8 Flash each rated all 90 clips separately, with one clip and the original prompt per request. Fresh contexts omitted the rewritten prompt, condition label, and source filename. Clips used a fixed shuffled order. Each judge scored five dimensions from 1 to 10: prompt alignment, audiovisual consistency, audio quality, video quality, and completeness/coherence. The shared rubric anchored 1 at absent or unusable, 3 at major failures, 5 at mixed success with substantial flaws, 7 at mostly successful with noticeable flaws, 9 at excellent with minor flaws, and 10 at fully successful with no apparent flaws. Intermediate integers were permitted. The overall score equally weighted the five ratings.

Huang et al.~\cite{huang2026vaj} introduced VA-Judger for joint audiovisual evaluation. We adapted its scoring instructions to rate a single clip, with 2~fps video sampling and greedy decoding. Gemini received the MP4 with audio, using the provider's default video processing and decoding settings~\cite{geminivideo}. In text evaluation, Chiang and Lee~\cite{chiang2023closer} found improved agreement with human ratings when judges explained their scores. Here, the rubric requested audiovisual evidence before the scores. Incomplete responses were retried. We fixed the rubric, equal weighting, and primary comparison before collecting these ratings.

\paragraph{Statistics.}
We focused on whether expanded prompts received higher overall scores than format-only prompts. We also compared both with raw prompts and explored differences in the individual rating dimensions. Each analysis matched the same original prompt and generation seed across conditions and included all 30 prompts, including zero score differences. We calculated paired mean changes and descriptive 95\% intervals from 10,000 bootstrap resamples of prompts (seed~42, percentile intervals). Intervals were unadjusted for multiple comparisons. The judges used the same clips, so their results were not independent replications. These intervals do not capture variation across generation seeds, writers, or evaluators.

\section{Results}

All clips received complete ratings from each judge. Expansion increased the overall score over formatting by 0.63 points under VA-Judger (95\% interval: $[+0.11, +1.15]$) and 0.15 under Gemini (95\% interval: $[-0.03, +0.35]$). Gemini's interval included zero. The mean overall scores were 8.95 versus 8.32 for VA-Judger and 6.76 versus 6.61 for Gemini.

\begin{figure}[H]
  \centering
  \includegraphics[width=0.95\textwidth]{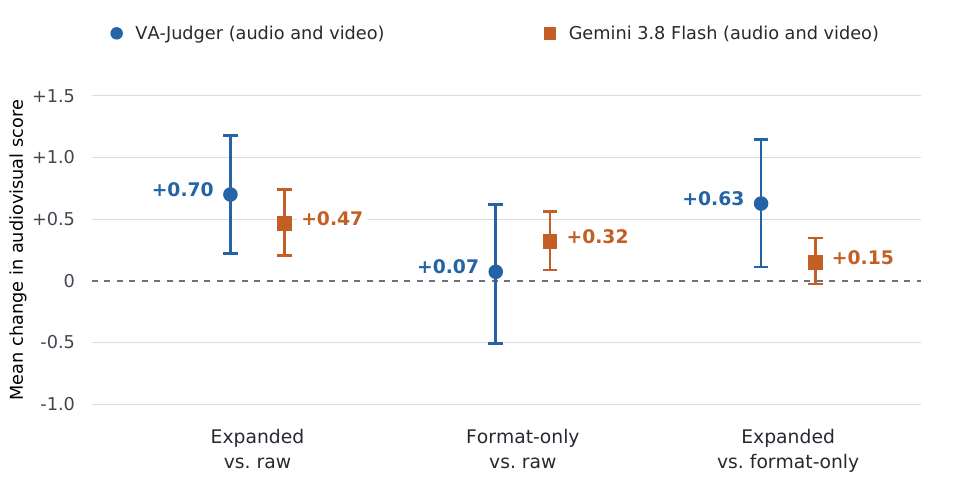}
  \caption{Paired mean changes in overall score across all 30 prompts. Positive changes favor the first condition named. Whiskers show descriptive 95\% paired-bootstrap intervals. The dashed line marks zero change. Ratings are not calibrated between judges.}
  \label{fig:results}
\end{figure}

Compared with raw prompts, expansion increased overall scores by 0.70 points for VA-Judger (95\% interval: $[+0.22, +1.18]$) and 0.47 for Gemini (95\% interval: $[+0.21, +0.74]$). Formatting alone increased scores by 0.07 for VA-Judger (95\% interval: $[-0.51, +0.62]$) and 0.32 for Gemini (95\% interval: $[+0.09, +0.56]$). Figure~\ref{fig:results} shows these paired mean changes. Table~\ref{tab:dimensions} gives the primary comparison's dimension changes.

\begin{table}[H]
\centering
\small
\begin{tabular}{@{}lrr@{}}
\hline
Dimension & VA-Judger & Gemini 3.8 Flash \\
\hline
Prompt alignment & $+0.97\;[+0.23,+1.73]$ & $+0.50\;[+0.13,+0.93]$ \\
Audiovisual consistency & $+0.50\;[-0.07,+1.10]$ & $+0.10\;[-0.13,+0.37]$ \\
Audio quality & $+1.00\;[+0.27,+1.80]$ & $+0.17\;[-0.07,+0.43]$ \\
Video quality & $+0.07\;[-0.43,+0.53]$ & $-0.03\;[-0.27,+0.20]$ \\
Completeness/coherence & $+0.60\;[0.00,+1.23]$ & $0.00\;[-0.27,+0.33]$ \\
\hline
\end{tabular}
\caption{Paired mean differences between expanded and format-only prompts on each 1--10 dimension. Positive values favor expansion. Brackets give descriptive 95\% paired-bootstrap intervals. Every dimension uses all 30 prompts.}
\label{tab:dimensions}
\end{table}

\section{Discussion}

The clearest shared gain beyond formatting was prompt alignment: +0.97 points for VA-Judger and +0.50 for Gemini (Table~\ref{tab:dimensions}). VA-Judger also showed an audio-quality gain of 1.00 point, compared with a video-quality change of 0.07. Its audio-quality gain exceeded its video-quality gain by 0.93 points (95\% interval: $[+0.13, +1.77]$). Visual-quality ratings changed little under either judge. The evidence therefore supports better adherence to the prompt, with an additional audio-quality gain under VA-Judger. The overall improvement beyond formatting remained uncertain under Gemini.

The audio difference has more than one plausible explanation. Expansion supplied detailed ambience and action-linked sound, but it also specified background music in 22 prompts, compared with one in the formatting control. Adding background music may have contributed to VA-Judger's higher audio-quality ratings. The experiment cannot separate that effect from better sound design or coordination with visible action. Audiovisual-consistency and completeness/coherence intervals included or touched zero under both judges, so improved synchronization or story coherence is not established as the cause of the score changes.

The experiment evaluated the combined effect of added content, greater prompt length, and occasional research. Evaluator behavior may also contribute to the different overall changes. Huang et al.~\cite{huang2026vaj} evaluated VA-Judger with pairwise comparisons, whereas this study used individual-clip ratings. Google's video-understanding documentation notes that default frame sampling can miss short visual events~\cite{geminivideo}. The 30 authored prompts and one generation seed support a controlled pilot of short clips. Generalization to new prompts, seeds, and longer videos remains untested.

\section*{AI assistance disclosure}

OpenAI Codex assisted with manuscript drafting and revision, analysis code, and figure preparation. The authors take responsibility for the final content.

\end{document}